\documentclass[times, review, 10pt]{elsarticle}

\usepackage{amssymb}
\usepackage{amsmath}
\usepackage{amsthm}

\usepackage{lineno}

\usepackage{multirow}
\usepackage{booktabs}
\usepackage{color}
\usepackage[table]{xcolor}
\usepackage[normalem]{ulem}
\usepackage[utf8]{inputenc}
\usepackage{hyperref}

\newtheorem{theorem}{Theorem}

\newdefinition{rmk}{Remark}
\newproof{pf}{Proof}
\newproof{pot}{Proof of Theorem \ref{thm}}

\newtheorem{assumption}{Assumption}
\newtheorem{corollary}{Corollary}

\begin{document}

\begin{frontmatter}



\title {A Novel Federated Learning Framework for Defending Against Gradient Inversion Attacks}


\author[1]{Pengxin Guo\fnref{fn1}} 
\author[2]{Ziying Huang\fnref{fn1}}
\author[3]{Shuang Zeng}
\author[4]{Xiaodan Zhang}
\author[5,6]{Weihong Ren}
\author[7]{Yuyin Zhou}
\author[1]{Liangqiong Qu\corref{cor1}}
\ead{liangqqu@hku.hk}

\affiliation[1]{organization={School of Computing and Data Science, The University of Hong Kong},
            city={Hong Kong},
            postcode={999077}, 
            country={China}}

\affiliation[2]{organization={Department of Electrical and Electronic Engineering, Southern University of Science and Technology},
            city={Shenzhen},
            postcode={518055}, 
            country={China}}

\affiliation[3]{organization={Department of Mathematics, The University of Hong Kong},
            city={Hong Kong},
            postcode={999077}, 
            country={China}}

\affiliation[4]{organization={College of Computer Science, Beijing University of Technology},
            city={Beijing},
            postcode={100124}, 
            country={China}}

\affiliation[5]{organization={State Key Laboratory of Robotics and Systems, School of Mechanical Engineering and Automation, Harbin Institute of Technology},
            city={Shenzhen},
            postcode={518055}, 
            country={China}}

\affiliation[6]{organization={State Key Laboratory of Robotics, Shenyang Institute of Automation, Chinese Academy of Sciences},
            city={Shenyang},
            postcode={110016}, 
            country={China}}

\affiliation[7]{organization={Department of Computer Science and Engineering, University of California},
            city={Santa Cruz, CA},
            postcode={95064}, 
            country={USA}}

\cortext[cor1]{Corresponding author}
\fntext[fn1]{Contribute equally.}

\begin{abstract}
Federated Learning (FL) offers a paradigm for collaborative machine learning model training among clients, preserving data privacy by avoiding direct data sharing. However, the gradient information exchanged in FL has been shown to be vulnerable to Gradient Inversion Attacks (GIA), which can reconstruct sensitive user data. To counter these threats, various privacy-enhancing technologies like Secure Multi-party Computing (SMC), Homomorphic Encryption (HE), and Differential Privacy (DP) have been integrated into FL. While effective, these methods often introduce significant trade-offs between privacy and model utility. By revisiting the key to privacy exposure in FL under GIA, which lies in the frequent sharing of model gradients that contain private data, we take a new perspective by designing a novel privacy-preserving FL framework that effectively ``breaks the direct connection'' between the shared parameters and the local private data to defend against GIA. Our proposed framework, HyperFL++, leverages hypernetworks to generate local model parameters, where only the hypernetwork parameters are aggregated by the server. To ensure practicality, HyperFL++ incorporates a Row-wise Kronecker Decomposition (RKD) to mitigate the memory and communication overhead from the hypernetwork. Comprehensive theoretical analysis and extensive experiments validate our method's effectiveness in protecting data privacy while maintaining high model utility. Our code is available at \href{https://github.com/Pengxin-Guo/HyperFL/tree/hyperfl%2B%2B}{https://github.com/Pengxin-Guo/HyperFL++}.
\end{abstract}



\begin{keyword}



Federated Learning \sep Gradient Inversion Attacks \sep Privacy Protection \sep Hypernetworks \sep Kronecker Decomposition

\end{keyword}

\end{frontmatter}




\section{Introduction}\label{sec:intro}

Deep Neural Networks (DNNs) have demonstrated significant success in various computer vision tasks, primarily due to access to extensive training datasets \cite{dosovitskiy2021image, wang2023unsupervised, zhang2026bridging}. However, in many real-world scenarios, training data is often distributed among different institutions, and sharing this data is commonly restricted by privacy and regulatory issues. To address these challenges, Federated Learning (FL) \cite{mcmahan2017communication, zeng2024tackling,guo2025exploringllm,zheng2025fedvlmbench,guo2026fed}, which utilizes collaborative and decentralized training of AI models across multiple clients while keeping personal data private, has emerged as an effective solution. In the FL paradigm, during each training round, clients perform local model updates using the latest copy of the model received from the server. Then, they send the updates to the server for aggregation. The server aggregates these updates to update the global model, which is then broadcasted to all clients for further training. By enabling collaborative training without raw data sharing, FL ensures data privacy protection.

\begin{figure}
  \centering
   \includegraphics[width=0.75\linewidth]{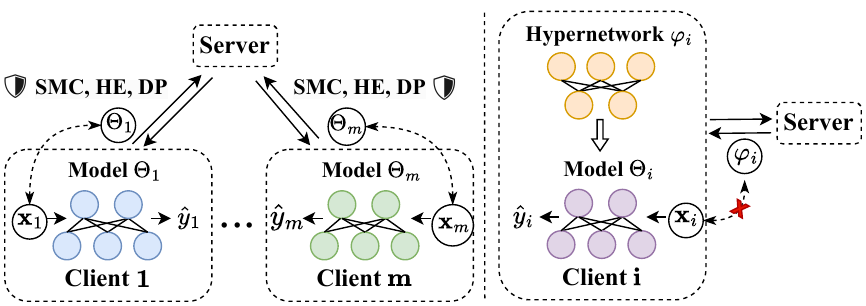}
   \caption{\textit{Left}. Current approaches focus on protective techniques for shared gradients, such as SMC, HE, and DP, which naturally entail significant privacy-utility trade-offs. \textit{Right}. A new FL framework is introduced that “breaks the direct connection” between the shared parameters and the local private data, aiming to achieve an optimal balance between privacy and utility.}
\label{fig:motivation}
\end{figure}

Despite its privacy-centric design, FL is vulnerable to Gradient Inversion Attacks (GIA) \cite{zhu2019deep, geiping2020inverting, huang2021evaluating, guo2026exploring}, which can reconstruct private training data from shared gradients. To mitigate this, defense mechanisms like Secure Multi-party Computing (SMC) \cite{yao1982protocols, bonawitz2017practical}, Homomorphic Encryption (HE) \cite{gentry2009fully, zhang2020privacy}, and Differential Privacy (DP) \cite{dwork2006differential, geyer2017differentially,shen2023share,zhang2026adaptive} have been adopted. As shown in Figure \ref{fig:motivation} (left), these methods apply protections to the exchanged gradients without altering the fundamental FL framework. However, they introduce a significant privacy-utility dilemma.

For example, SMC and HE offer strong security but impose prohibitive computational and communication overheads, especially for large-scale DNNs \cite{bonawitz2017practical, zhang2020privacy}. Conversely, DP is more lightweight but often compromises model accuracy to achieve meaningful privacy guarantees \cite{geyer2017differentially, mcmahan2018learning}. These limitations motivate us to explore a different path, leading to our central research question:
\begin{center}
    \textit{\quad Can we develop a new FL framework that provides a favorable privacy-utility trade-off against GIA without depending on current defense mechanisms?}
\end{center}

To achieve this, we revisit the key to privacy exposure in FL under GIA, which stems from the frequent sharing of model gradients containing private information. Most existing efforts focus on developing advanced defense mechanisms for the shared gradients to improve privacy in FL, as depicted in the left section of Figure \ref{fig:motivation}. In contrast, our approach seeks to create a novel privacy-preserving FL framework that ``breaks the direct connection'' between shared parameters and local private data to guard against GIA.
To accomplish this, we investigate the potential of hypernetworks—a type of deep neural network that generates weights for another network \cite{david2017hypernetworks}—as a promising solution, as shown in the right section of Figure \ref{fig:motivation} and Figure \ref{fig:framework}.

To this end, we present \textbf{HyperFL++}, an innovative FL framework that employs a dual approach—network decomposition and hypernetwork sharing— to ``break the direct connection'' between shared parameters and local private data, as illustrated in Figure~\ref{fig:framework}.
Inspired by findings that FL models exhibit greater diversity in their final layers \cite{xu2023personalized, shen2023share}, we partition each local model into a shared feature extractor and a private classifier. This design improves performance in heterogeneous settings and confines some data-sensitive components locally. To further enhance privacy, an auxiliary hypernetwork generates the feature extractor's parameters from a private, client-specific embedding. Instead of sharing the feature extractor's gradients, clients only transmit the hypernetwork's parameters to the server for aggregation. This auxiliary hypernetwork sharing approach effectively ``breaks the direct connection'' between shared parameters and local private data, preserving privacy while facilitating interaction and information exchange among clients. To make HyperFL++ efficient, we introduce a \textbf{Row-wise Kronecker Decomposition (RKD)} technique that compresses the hypernetwork's parameters, significantly reducing the associated memory and communication costs, as detailed in Figure \ref{fig:framework} (right). 
This operation largely reduces the parameters associated with the hypernetwork, minimizing the communication overhead introduced by hypernetworks.

The HyperFL++ framework is designed to be both flexible and scalable. We propose two main configurations: (1) the Main Configuration, which generates the entire feature extractor and is suited for smaller networks, and (2) HyperFL++ for Large Pre-trained Models (HyperFL++-LPM), which uses a hypernetwork to generate parameters for lightweight adapters \cite{houlsby2019parameter} to fine-tune large, fixed pre-trained models for more complex tasks (see Figure \ref{fig:framework_v2}).
Our theoretical and empirical results confirm that HyperFL++ robustly defends against GIA while delivering model performance and convergence rates comparable to the standard FedAvg \cite{mcmahan2017communication}. We hope this work encourages a shift in focus toward designing inherently private FL frameworks rather than relying solely on external defense mechanisms.

Our main contributions are:
\begin{itemize}
    \item We propose HyperFL++, a novel privacy-preserving FL framework that defends against GIA by ``breaking the direct connection'' between shared parameters and local data, offering a new perspective on FL privacy.
    \item We introduce a Row-wise Kronecker Decomposition (RKD) method to make the hypernetwork-based approach practical by minimizing its communication and memory footprint.
    \item We present two key configurations of HyperFL++: (1) the Main Configuration, which is designed for simpler tasks with smaller networks and learns the full parameters of the feature extractor directly; (2) HyperFL++-LPM, which is aimed at complex tasks by utilizing pre-trained models as fixed feature extractors while generating trainable adapter parameters through a hypernetwork.
    \item Our theoretical analysis and extensive experimental results show that HyperFL++ provides strong privacy protection against GIA while achieving competitive model utility and convergence.
\end{itemize}

\section{Related Work}
\label{sec:related_work}

\subsection{Gradient Inversion Attacks} 

Gradient Inversion Attacks (GIA) are a class of privacy attacks that exploit the gradients shared in collaborative learning to reconstruct sensitive training data \cite{zhu2019deep}. The underlying principle is that gradients, which guide model updates, inherently contain information about the data that produced them. The fundamental concept of GIA involves deliberately altering the dummy input to maximize the gradients related to the sensitive information of interest. By iteratively modifying the dummy input based on these gradients, an attacker can progressively approximate sensitive details, such as private attributes or specific training data samples that the model was trained on \cite{phong2017privacy, zhu2019deep, geiping2020inverting, yin2021see}.

Early work by Phong et al. demonstrated the feasibility of GIA on shallow networks \cite{phong2017privacy}. Subsequent research extended these attacks to deeper models. For instance, Zhu et al. introduced a method combining pseudo-label estimation and optimization to enhance image reconstruction quality \cite{zhu2019deep}. A significant advancement came from Geiping et al., who replaced the traditional L2 distance loss with cosine similarity, dramatically improving the fidelity of reconstructed images, even for complex datasets like ImageNet \cite{geiping2020inverting}. The continued success of these attacks highlights that shared gradients pose a considerable privacy threat, underscoring the need for robust defense mechanisms to prevent information leakage.

\subsection{Privacy Protection in Federated Learning} 

While FL is designed to keep data localized, the necessity of sharing model updates opens an attack surface for GIA \cite{geiping2020inverting}. Consequently, researchers have integrated various privacy-preserving technologies into FL. These can be broadly categorized into three groups: cryptographic methods like Secure Multi-party Computing (SMC) \cite{yao1982protocols, bonawitz2017practical} and Homomorphic Encryption (HE) \cite{gentry2009fully, zhang2020privacy}, and perturbation-based methods like Differential Privacy (DP) \cite{dwork2006differential, geyer2017differentially,shen2023share}.
Cryptographic solutions offer strong security guarantees by performing computations on encrypted data, but their high computational and communication overheads make them impractical for many large-scale deep learning applications \cite{bonawitz2017practical, zhang2020privacy}. DP-based methods, which involve adding calibrated noise to gradients, are more lightweight but often lead to a degradation in model performance, creating a difficult trade-off between privacy and utility \cite{geyer2017differentially, mcmahan2018learning}.

Beyond these theoretically-grounded defenses, several empirical strategies have been proposed. These include gradient pruning or masking \cite{zhu2019deep, huang2021evaluating}, which zeroes out small-magnitude gradients, and unstructured noise addition \cite{zhu2019deep}. Other notable methods are Soteria \cite{sun2020provable}, which perturbs intermediate feature representations, and PRECODE \cite{scheliga2022precode}, which uses a variational module to obscure latent features. However, as demonstrated in their respective studies, these empirical methods also struggle with the inherent privacy-utility trade-off \cite{huang2021evaluating, scheliga2022precode, sun2020provable}. Our work diverges from these approaches by proposing a new FL framework that redesigns the information-sharing mechanism itself, using hypernetworks to create a more favorable privacy-utility balance.

\subsection{Hypernetworks in Federated Learning} 

A hypernetwork is a neural network that generates the weights for a separate, primary network, known as the target network \cite{david2017hypernetworks,guo2021deep}. This concept has recently been applied to FL, primarily to address statistical heterogeneity by generating personalized models for each client \cite{shamsian2021personalized, tashakori2023semipfl, lin2023federated}.
For example, pFedHN \cite{shamsian2021personalized} uses a central hypernetwork to generate client-specific parameters from learned client embeddings, enabling efficient personalization. Similarly, SemiPFL \cite{tashakori2023semipfl} employs a global hypernetwork to create personalized autoencoders for clients in a semi-supervised setting. A common thread in these methods is that the hypernetwork and client embeddings are managed by the central server. This design, while effective for personalization, gives the server complete knowledge of the local models, making it vulnerable to GIA, as our experiments confirm (see Table \ref{tab:attack} and Figure \ref{fig:attack}).

Our work, HyperFL++, takes a different approach. Instead of focusing on personalization, we leverage the hypernetwork architecture as a privacy-preserving mechanism. By carefully designing what is shared (only the hypernetwork parameters) and what remains private (the classifier and client embeddings), we fundamentally alter the information available to the server, thereby minimizing the risk of data leakage while maintaining high model utility.

\section{Methodology}
\label{sec:method}

In this section, we first formalize the problem setting of federated learning and then introduce our proposed HyperFL++ framework in detail.

\subsection{Problem Formulation}

\begin{figure*}
  \centering
   \includegraphics[width=0.99\linewidth]{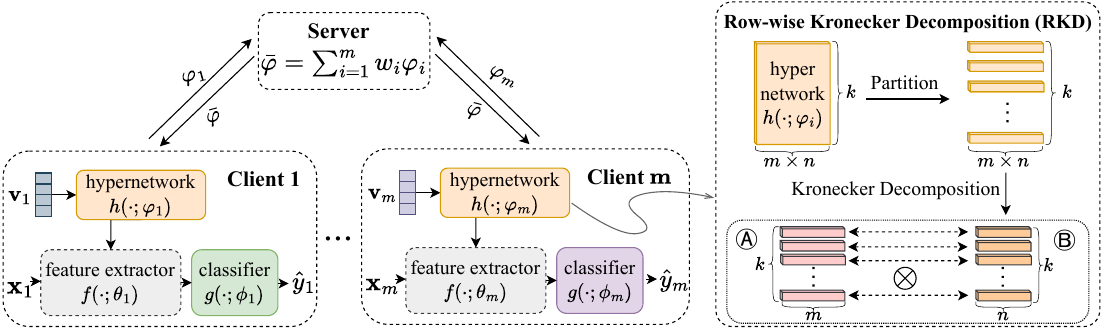}
   \caption{\textit{Left)}. Illustration of \textbf{HyperFL++}. HyperFL++ divides each client's network into two components: the feature extractor $f(; \theta_i)$ and the classifier head $g(;{\phi _i})$. An auxiliary hypernetwork, denoted as $h(;{\varphi _i})$, is employed to create the feature extractor $f(; \theta_i)$  using the client’s specific private embedding vector $\mathbf{v}_i$, such that ${\theta _i} = h({{\bf{v}}_i};{\varphi _i})$. These parameters are then utilized to derive features from the input ${x}_i$, which are subsequently processed by the classifier to produce the output $\hat{y}_i$, represented as $\hat{y}_i = g( f({x}_i; \theta_i); \phi_i)$. During the FL process, the hypernetwork $\varphi_i$ is the only element shared with the server, while all other components remain private. This approach effectively addresses privacy leakage concerns. \textit{Right).} Row-wise Kronecker Decomposition (\textbf{RKD}). The hypernetwork parameters (matrices) are first partitioned into rows, and then Kronecker decomposition is applied to each row. $\otimes$ denotes Kronecker product.}   
\label{fig:framework}
\end{figure*}

Consider a federated learning system with $m$ clients and a central server. The clients aim to collaboratively train their models without sharing their raw private data. Each client $i$ possesses a local data distribution $P_{XY}^{(i)}$ over the input space $\mathcal{X}$ and label space $\mathcal{Y}$, where there are $K$ classes in total.
The local model for client $i$ is denoted by $\Theta_i$, and the associated loss function is $\ell: \mathcal{X} \times \mathcal{Y} \rightarrow \mathbb{R}_+$. The global optimization objective in FL is to find the set of model parameters $\Theta = \{\Theta_1, \ldots, \Theta_m\}$ that minimizes the average loss across all clients:
\begin{equation} \label{eq:fl_obj_1}
    \arg \min _{\Theta} \frac{1}{m} \sum_{i=1}^m \mathbb{E}_{({x}, y) \sim P_{X Y}^{(i)}}\left[\ell\left(\Theta_i ; {x}, y\right)\right].
\end{equation}
In the standard FedAvg algorithm \cite{mcmahan2017communication}, all clients share identical model parameters, i.e., $\Theta_1 = \cdots = \Theta_m$. In contrast, personalized FL allows for client-specific parameters, where $\Theta_i$ can differ across clients to better handle data heterogeneity \cite{xu2023personalized,guo2025selective}.

Since the true data distributions are unknown, the objective in Eq. (\ref{eq:fl_obj_1}) is typically approximated using Empirical Risk Minimization (ERM). Each client $i$ has a local dataset $\mathcal{D}_i=\left\{\left({x}_i^l, y_i^l\right)\right\}_{l=1}^{n_i}$ of $n_i$ IID samples drawn from $P_{XY}^{(i)}$. The training objective thus becomes minimizing the empirical risk:
\begin{equation} \label{eq:fl_obj_2}
    \arg \min _{\Theta} \frac{1}{m} \sum_{i=1}^m \mathcal{L}_i(\Theta_i),
\end{equation}
where $\mathcal{L}_i(\Theta_i) = \frac{1}{n_i} \sum_{l=1}^{n_i} \ell(\Theta_i; {x}_i^l, y_i^l)$ is the average loss computed on the local training data of client $i$.

\subsection{Main Configuration of HyperFL++}

This section details the complete workflow of the \textbf{HyperFL++} framework, followed by an explanation of the low-rank factorization technique applied to the hypernetwork.

\subsubsection{HyperFL++ Framework}

As depicted in Figure \ref{fig:framework}, in the Main Configuration of HyperFL++, each client $i$'s classification network, defined by $\Theta_i = \{\theta_i, \phi_i\}$, is composed of a feature extractor $f: \mathcal{X} \rightarrow \mathbb{R}^d$ with parameters $\theta_i$ and a classifier $g: \mathbb{R}^d \rightarrow \mathbb{R}^K$ with parameters $\phi_i$. Here, $d$ is the feature dimension and $K$ is the number of classes. Each client also has a private embedding $\mathbf{v}_i$ and a hypernetwork $h$ parameterized by $\varphi_i$. The hypernetwork generates the parameters for the feature extractor, i.e., $\theta_i = h(\mathbf{v}_i; \varphi_i)$. This design enables the creation of personalized feature extractors based on unique client embeddings, which can be either fixed or trainable; in this work, we use trainable vectors. The optimization objective from Eq. (\ref{eq:fl_obj_2}) is then reformulated as:
\begin{equation} \label{eq:fl_obj_ours}
    \arg \min _{\varphi, \phi, \mathbf{v}} \frac{1}{m} \sum_{i=1}^m \mathcal{L}_i(h(\mathbf{v}_i; \varphi_i), \phi_i),
\end{equation}
where $\varphi = \{\varphi_1, \ldots, \varphi_m\}$, $\phi = \{\phi_1, \ldots, \phi_m\}$, and $\mathbf{v} = \{\mathbf{v}_1, \ldots, \mathbf{v}_m\}$. Note that the feature extractor parameters are generated by the hypernetwork and are not directly trained, whereas the client embeddings, hypernetwork parameters, and classifier parameters are all trainable.

To ``break the direct connection'' between shared parameters and private local data, thereby defending against GIA while maintaining high performance, clients only upload their hypernetwork parameters to the server for aggregation. The classifier and the private client embedding remain local. As shown in Figure \ref{fig:framework}, in each communication round, the server broadcasts the aggregated hypernetwork parameters $\bar{\varphi}$ to the clients. Each client $i$ updates its local hypernetwork parameters $\varphi_i$ with $\bar{\varphi}$ and proceeds with local training. After local training, each client sends its updated $\varphi_i$ back to the server for aggregation, while keeping its classifier parameters $\phi_i$ and client embedding $\mathbf{v}_i$ private. The server then aggregates the received $\varphi_i$ to compute the new global $\bar{\varphi}$ for the next round. This process facilitates information sharing among clients while preserving privacy.

\paragraph{Local Training Procedure} During each local training phase, a client first updates its hypernetwork parameters $\varphi_i$ with the aggregated parameters $\bar{\varphi}$ from the server. Then, it performs iterative updates using stochastic gradient descent as follows:
\begin{itemize}
    \item \textbf{Step 1: Update Classifier $\phi_i$}. With $\varphi_i$ and $\mathbf{v}_i$ fixed, the classifier parameters $\phi_i$ are trained for one epoch:
    \begin{equation} \label{eq:update_classifier}
    \phi_i \leftarrow \phi_i - \eta_g \nabla_{\phi_i} \ell\left(h(\mathbf{v_i}; \varphi_i), \phi_i ; \xi_i\right),
    \end{equation}
    where $\xi_i$ is a mini-batch of data and $\eta_g$ is the learning rate for the classifier.
    
    \item \textbf{Step 2: Update Hypernetwork $\varphi_i$ and Embedding $\mathbf{v}_i$}. After updating the classifier, the client updates the hypernetwork parameters $\varphi_i$ and the client embedding $\mathbf{v}_i$ for multiple epochs with the new $\phi_i$ fixed:
    \begin{align} \label{eq:update_hypernetwork}
    &\varphi_i \leftarrow \varphi_i - \eta_h \nabla_{\varphi_i} \ell\left(h(\mathbf{v_i}; \varphi_i), \phi_i ; \xi_i\right) \nonumber \\
    &\mathbf{v}_i \leftarrow \mathbf{v}_i - \eta_v \nabla_{\mathbf{v}_i} \ell\left(h(\mathbf{v_i}; \varphi_i), \phi_i ; \xi_i\right),
    \end{align}
    where $\eta_h$ and $\eta_v$ are the learning rates for the hypernetwork and the client embedding, respectively.
\end{itemize}

\paragraph{Global Aggregation} Following standard FL protocols like FedAvg \cite{mcmahan2017communication}, the server aggregates the hypernetwork parameters from all clients via weighted averaging:
\begin{equation} \label{eq:server_agg}
\bar{\varphi} = \sum_{i=1}^m w_i \varphi_i,
\end{equation}
where $w_i = \frac{n_i}{\sum_{j=1}^m n_j}$ is the aggregation weight for client $i$, typically proportional to its local data size.

However, a key challenge in HyperFL++ is the increased communication overhead, as the hypernetwork's parameters are generally larger than those of the feature extractor it generates \cite{david2017hypernetworks}. For instance, if a convolutional layer's feature extractor parameters are $\theta_l \in \mathbb{R}^{\text{IC} \times \text{OC} \times \text{KS} \times \text{KS}}$, the hypernetwork generating them would have parameters $\varphi_l \in \mathbb{R}^{k \times (\text{IC} \times \text{OC} \times \text{KS} \times \text{KS})}$, where $k$ is the hypernetwork's hidden dimension. This results in a $k$-fold increase in parameter size. To address this and make HyperFL++ practical, we introduce a low-rank factorization method to compress the hypernetwork and reduce its computational and memory footprint.

\subsubsection{Row-wise Kronecker Decomposition (RKD)}

A primary challenge in employing hypernetworks is their potential size. The hypernetwork must generate all the parameters for the target network's feature extractor, which can lead to a substantial number of parameters within the hypernetwork itself. This increase in model size translates directly to higher communication overhead and memory requirements, potentially undermining the practicality of the framework. To overcome this critical bottleneck and make HyperFL++ efficient, we introduce a powerful compression technique: \textbf{Row-wise Kronecker Decomposition (RKD)}.

RKD is a specialized low-rank factorization method designed to compress the large parameter matrices of the hypernetwork. The core idea is to represent a large matrix as a composition of much smaller ones, thereby reducing the number of parameters that need to be stored and communicated. As illustrated in the right panel of Figure \ref{fig:framework}, the process works as follows:

Let the parameters for a single layer of the hypernetwork be represented by a large matrix $\varphi \in \mathbb{R}^{k \times N}$, where $k$ is the dimension of the hypernetwork's hidden layer and $N$ is the total number of parameters in the target feature extractor layer (e.g., $N = \text{IC} \times \text{OC} \times \text{KS} \times \text{KS}$ for a convolutional layer). Instead of storing the full $\varphi$ matrix, we decompose it into two smaller factor matrices, $A \in \mathbb{R}^{k \times m}$ and $B \in \mathbb{R}^{k \times n}$, where the dimensions are chosen such that $m \times n = N$.

The ``row-wise'' aspect of RKD means we reconstruct each row of the original matrix $\varphi$ independently. Specifically, the $r$-th row of $\varphi$ is reconstructed by taking the Kronecker product of the $r$-th row of $A$ and the $r$-th row of $B$:
\begin{equation}
    \varphi_{r,:} =  A_{r,:} \otimes B_{r,:}, \quad \text{for } r = 1, \ldots, k,
\end{equation}
where $\otimes$ denotes the Kronecker product \cite{van2000ubiquitous}. During training, only the factor matrices $A$ and $B$ are stored and updated.

The compression benefit of this decomposition is significant. The original number of parameters to store or communicate for the matrix $\varphi$ is $k \times N = k \times mn$. With RKD, we only need to handle the parameters of $A$ and $B$, which total $k \times m + k \times n = k \times (m+n)$. By choosing the factors $m$ and $n$ to be approximately equal (i.e., $m \approx n \approx \sqrt{N}$), the parameter complexity is dramatically reduced from $O(kN)$ to $O(k\sqrt{N})$. This substantial compression makes the hypernetwork-based approach computationally feasible and communication-efficient, enabling HyperFL++ to achieve a favorable balance between privacy, utility, and system performance.

\subsection{HyperFL++ for Large Pre-trained Models (HyperFL++-LPM)}

\begin{figure}[h]
  \centering
   \includegraphics[width=0.6\linewidth]{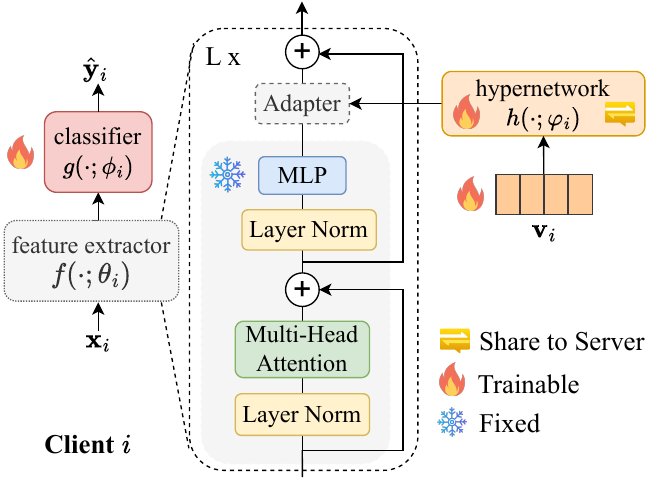}
   \caption{The \textbf{HyperFL++-LPM} framework in each client. In this framework, the weights of the pre-trained model remain fixed, while only the classifier, hypernetwork, and client embedding are trainable. It’s important to note that $\theta$ refers to the parameters of the adapters.}
\label{fig:framework_v2}
\end{figure}

While the main configuration of HyperFL++ is effective for moderately sized networks, generating the entire feature extractor for very large-scale models, such as Vision Transformers (ViTs) \cite{dosovitskiy2021image} or other foundation models, can become impractical due to the sheer number of parameters. The hypernetwork required to generate such a large target network would itself be enormous, leading to prohibitive computational and memory costs.

To address this scalability challenge and extend the applicability of our framework to state-of-the-art large models, we introduce an adaptation named \textbf{HyperFL++-LPM} (for Large Pre-trained Models). This configuration, illustrated in Figure \ref{fig:framework_v2}, leverages the power of pre-trained models by integrating them with parameter-efficient fine-tuning (PEFT) techniques, such as adapters \cite{houlsby2019parameter}.

The core idea behind HyperFL++-LPM is to keep the vast majority of the pre-trained model's parameters frozen and only fine-tune a small set of additional parameters. Instead of generating the entire feature extractor, the hypernetwork's role is shifted to generating only the parameters for these lightweight adapter modules. These adapters are small neural network layers inserted within the architecture of the larger pre-trained model. By training only these adapters, we can adapt the powerful, general-purpose features of the large model to specific downstream tasks and client data distributions without incurring the cost of full model fine-tuning.

In this setup, the client model consists of a large, fixed pre-trained feature extractor, the trainable adapters whose parameters are generated by the hypernetwork, and a private classifier. The training process remains similar to the main configuration: clients update their local hypernetwork, client embedding, and classifier, but the updates to the feature extractor are now confined to the adapter layers. This approach allows HyperFL++ to effectively harness the capabilities of powerful pre-trained models for complex tasks while maintaining the core privacy and efficiency benefits of our framework. It provides a practical and scalable solution for applying privacy-preserving federated learning in the era of large foundation models.

\section{Analysis and Insights}
\label{sec:theo_ana}

\subsection{Privacy Analysis} \label{sec:any_privacy}

In this section, we conduct a detailed privacy analysis of the HyperFL++ framework. We consider the common scenario where the server is an \textit{honest-but-curious} adversary, meaning it adheres to the training protocol but attempts to infer private client data from the shared model updates \cite{zhu2019deep,geiping2020inverting,yin2021see,guo2026exploring}.

\subsubsection{Background on Gradient Inversion Attacks}
As established by prior research \cite{phong2017privacy, zhu2019deep, geiping2020inverting}, it is possible to reconstruct input data from gradients. This is achieved by solving an optimization problem. Given a model with parameters $\Theta$ and the gradients $\nabla_{\Theta} \mathcal{L}_{\Theta} (x^*, y^*)$ computed from a private data batch $(x^*, y^*)$, an attacker aims to find a reconstructed input $x$ that approximates $x^*$:
\begin{equation} \label{eq:attack}
    \arg \min _x \mathcal{L}_{\text {grad }}\left(x ; \Theta, \nabla_\Theta \mathcal{L}_\Theta\left(x^*, y^*\right)\right) + \alpha \mathcal{R}_{\text {aux }}(x).
\end{equation}
Here, $\mathcal{L}_{\text {grad}}$ is a loss function that measures the similarity between the gradients of the reconstructed data $x$ and the target gradients $\nabla_\Theta \mathcal{L}_\Theta(x^*, y^*)$. The term $\mathcal{R}_{\text {aux }}(x)$ is a regularization component that incorporates prior knowledge about natural images to improve reconstruction quality, with $\alpha$ as its weighting factor. Different GIA methods vary in their choice of the gradient loss and regularization terms \cite{zhu2019deep, geiping2020inverting, yin2021see}.

\subsubsection{Analysis of the Proposed Defense Mechanism} \label{sec:why_work}
In contrast to standard FL frameworks where clients upload gradients of the entire model, HyperFL++ requires clients to share only the hypernetwork parameters. Consequently, an attacker's objective function for reconstructing data $x$ must be adapted:
\begin{equation} \label{eq:att_ours}
    \arg \min _{x} \mathcal{L}_{\text {grad }}\left(x; \varphi, \nabla_\varphi \mathcal{L}_{\varphi, \phi}\left(x^*, y^*, \mathbf{v}^*\right)\right) + \alpha \mathcal{R}_{\text {aux }}(x),
\end{equation}
where $\varphi$ are the hypernetwork parameters, $\phi$ are the classifier parameters, and $\mathbf{v}^*$ is the private client embedding. The crucial difference between Eq. (\ref{eq:attack}) and Eq. (\ref{eq:att_ours}) is that the attacker only has access to the gradients of the hypernetwork, not the entire model.

To reconstruct the input $x$, the attacker must first gain information about the feature extractor, whose parameters are generated by the hypernetwork based on the client embedding. This turns the attack into a challenging bi-level optimization problem:
\begin{equation} \label{eq:att_ours_bilevel}
\begin{aligned}
&\arg \min _{x} \mathcal{L}_{\text {grad }}\left(x, \hat{\mathbf{v}}; \varphi, \Delta_\theta\right) + \alpha \mathcal{R}_{\text {aux }}(x)  \\
\mathrm{s.t.}~~&\hat{\mathbf{v}} = \arg \min _{\mathbf{v}} \mathcal{L}_{\text {grad }}\left(x, \mathbf{v}; \varphi, \nabla_\varphi \mathcal{L}_{\varphi, \phi}\left(x^*, y^*, \mathbf{v}^*\right)\right),
\end{aligned}
\end{equation}
where the lower-level problem aims to recover the client embedding $\mathbf{v}^*$. However, solving this is infeasible. According to the chain rule, computing the hypernetwork's gradient $\nabla_\varphi \mathcal{L}$ requires the classifier's gradient $\nabla_\phi \mathcal{L}$. Since the classifier is kept private on the client's device and never shared, the server cannot compute this gradient. This effectively breaks the inversion process.

Even if an attacker attempts to bypass the classifier by directly optimizing for the hypernetwork's output (the feature extractor parameters $\theta$), the problem remains intractable. Unlike inferring a low-dimensional label vector \cite{zhao2020idlg, yin2021see}, the output $\theta$ is a high-dimensional and complex set of parameters. Simultaneously optimizing for both the input data $x$ and the high-dimensional parameters $\theta$ creates an enormous search space, making a successful reconstruction practically impossible \cite{zhu2019deep, huang2021evaluating}.

In summary, HyperFL++'s architecture, which combines a hypernetwork with a private classifier and client embedding, creates a strong defense against GIA by fundamentally disrupting the information flow required for a successful attack. This is further validated by our experimental results in Section \ref{sec:exp_privacy}.

\subsection{Utility Analysis} \label{sec:utility_ana}

We now analyze the utility of the HyperFL++ framework, focusing on its convergence properties and generalization bound. Detailed proofs are available in the Appendix.

\subsubsection{Convergence Analysis} \label{sec:convergence_ana}
We adopt standard assumptions from the FL literature \cite{li2019convergence,guo2025selective} for smooth, non-convex optimization landscapes.

\begin{assumption} 
\label{assumption1}
(L-smoothness) The local loss functions $\mathcal{L}_1, \mathcal{L}_2, \cdots, \mathcal{L}_m$  are all L-smooth, meaning that for any pairs $(\phi_j, \varphi_j,\mathbf{v}_j)$ and $(\phi_k,\varphi_k,\mathbf{v}_k)$, the following holds:
$\mathcal{L}_i(h(\mathbf{v}_k,\varphi_k),\phi_k) \leq \notag \mathcal{L}_i(h(\mathbf{v}_j, \varphi_j),\phi_j) + ((\phi_k,\varphi_k,\mathbf{v}_k)-(\phi_j,\varphi_j,\mathbf{v}_j))\nabla \mathcal{L}_i(h(\mathbf{v}_j,\varphi_j), \phi_j) + \\ \frac{L}{2}\|(\phi_k,\varphi_k,\mathbf{v}_k)-(\phi_j,\varphi_j,\mathbf{v}_j)\|_2^2
$.
\end{assumption}

\begin{assumption}
\label{assumption2}
(Bounded Variance) Let $\xi_i^t$ be sampled uniformly at random from the local data of the $i$-th client at the $t$-th training step. The variance of the stochastic gradients for each client and each variable is bounded as follows: 

\begin{equation*}
\begin{aligned}
&\mathbb{E}\left\|\nabla_\phi L_i\left(h\left(\mathbf{v}_i^t, \varphi_i^t\right), \phi_i^t, \xi_i^t\right)-\nabla_\phi L_i\left(h\left(\mathbf{v}_i^t, \varphi_i^t\right), \phi_i^t\right)\right\|^2 \leq  \sigma_i^2, \\
&\mathbb{E}\left\|\nabla_\varphi L_i\left(h\left(\mathbf{v}_i^t, \varphi_i^t\right), \phi_i^{t+1}, \xi_i^t\right)-\nabla_\varphi L_i\left(h\left(\mathbf{v}_i^t, \varphi_i^t\right), \phi\right.\right. \left.\left._i^{t+1}\right)\right\|^2 \leq \sigma_i^2, \\
&\mathbb{E}\left\|\nabla_\mathbf{v} L_i\left(h\left(\mathbf{v}_i^t, \varphi_i^t\right), \phi_i^{t+1}, \xi_i^t\right)-\nabla_\mathbf{v} L_i\left(h\right.\right. \left.\left.\left(\mathbf{v}_i^t, \varphi_i^t\right), \phi_i^{t+1}\right)\right\|^2 \leq \sigma_i^2,
\end{aligned}
\end{equation*}
for $i=1, \cdots, m$.
\end{assumption}

\begin{assumption}
\label{assumption3}
(Bounded Gradient) The expected squared norm of the stochastic gradients is uniformly bounded, specifically: 
\begin{equation*}
\begin{aligned}
\mathbb{E}\left\|\nabla_\phi L_i\left(h\left(\mathbf{v}_i^t, \varphi_i^t\right),\right.\right. \left.\left. \phi_i^t, \xi_i^t\right)\right\|^2 &\leq G^2, \\ \mathbb{E}\left\|\nabla_\varphi L_i\left(h\left(\mathbf{v}_i^t, \varphi_i^t\right), \phi_i^{t+1}, \xi_i^t\right)\right\|^2 &\leq G^2, \\
\mathbb{E}\left\|\nabla_\mathbf{v} L_i\left(h\left(\mathbf{v}_i^t, \varphi_i^t\right), \phi_i^{t+1}, \xi_i^t\right)\right\|^2 &\leq G^2,
\end{aligned}
\end{equation*}
for all $i=1,\cdots, m$ and $t=0, \cdots, T-1$. Here, $T$ represents the total number of training steps for each client.
\end{assumption}

Under these assumptions, we establish the convergence rate for HyperFL++.
\begin{theorem}
\label{theorem1}
Assuming that Assumptions \ref{assumption1}, \ref{assumption2} and \ref{assumption3} are satisfied, and defining $L$, $M$, $\sigma_i$, $G$ as stated, let $\eta_{\min }=\min \left\{\eta_g, \frac{1}{2} \eta_h, \eta_v\right\}$ and $E$ represent the number of local training iterations between two communication rounds. Then we have:
\begin{equation}
\frac{1}{m T} \sum_{i=1}^m \sum_{t=1}^T \mathbb{E}\left[\left\|\nabla\mathcal{L}_i^t\right\|^2\right] \leq 2 \sqrt{\frac{L M G^2 D}{2 T}},
\end{equation}
where $\mathcal{L}_i^0 - \mathcal{L}_i^* \leq D, \forall i$, and
$
\eta_g^2+\eta_v^2+(E-1) \eta_h^2+\frac{E-1}{L} \eta_h \leq M \eta_{min}^2.
$
\end{theorem} 
Based on Theorem \ref{theorem1}, we achieve a convergence rate of $O(\frac{1}{\sqrt{T}})$ towards the stationary solution under smooth and non-convex conditions. This rate is comparable to that of FedAvg in non-convex situations \cite{yu2019parallel}. Additionally, we can accelerate convergence for Polyak-Lojasiewicz (PL) functions \cite{karimi2016linear}, which are frequently encountered in non-convex optimization problems.

\begin{assumption}
\label{pl_assumption}
(PL Condition) A function $f$ is referred to as a $\mu$-PL function if there exists a constant $\mu > 0$ such that it satisfies the condition:
\begin{equation*}
\|\nabla f(x)\|^2 \geq 2\mu\left(f(x) - \inf_{x^{\prime}}f(x^{\prime})\right), \quad \forall x.
\end{equation*}
We assume that all functions $\mathcal{L}_1, \cdots, \mathcal{L}_m$ are $\mu$-PL functions, and we denote $\inf_{x^{\prime}}f(x^{\prime})$ simply as $f^*$.
\end{assumption}

\begin{corollary}
\label{corollary1}
Given the assumptions along with $\eta_{min}$, $L$, $M$ and $D$ defined in Theorem \ref{theorem1}, and under the additional Assumption \ref{pl_assumption}, we have:
\begin{equation}
\begin{aligned}
\mathbb{E}\left[\frac{1}{m} \sum_{i=1}^m \mathcal{L}_i\left(h\left(\mathbf{v}_i^t ; \varphi_i^t\right), \phi_i^t\right)\right]-\mathcal{L}^* \leq \left(1-2\eta_{\min } \mu\right)^{t+1} D+\eta_{\min } \frac{L M G^2}{4 \mu}.
\end{aligned}
\end{equation}
If we set $\eta_{\min } \leq \frac{\mu \epsilon}{L M G^2}$, after $O\left(\frac{1}{\epsilon} \log \left(\frac{1}{\epsilon}\right)\right)$ steps, we have that:
\begin{equation}
\mathbb{E}\left[\frac{1}{m} \sum_{i=1}^m \mathcal{L}_i\left(h\left(\mathbf{v}_i^t ; \varphi_i^t\right), \phi_i^t\right)\right]-\mathcal{L}^* \leq \epsilon.
\end{equation}
\end{corollary}
When using PL functions, the convergence rate of HyperFL++ surpasses that obtained through smoothness assumptions alone.

\subsubsection{Generalization Bound} \label{sec:generalization_bound}

We also provide a generalization bound for HyperFL++ based on the framework from \cite{baxter2000model}.
\begin{assumption}
\label{assumption_gen_bound}
(Lipschitz Condition) We assume that the weights of hypernetworks $\varphi_i$, the client embeddings $\mathbf{v}_i$ and the classifier weights $\phi_i$ are bounded in a ball of radius $R$, in which the following Lipschitz conditions hold:
\begin{equation*}
\begin{aligned}
    &\left| \mathcal{L}_i(h(\varphi_i; \mathbf{v}_i), \phi_i)  - \mathcal{L}_i(h(\varphi_i; \mathbf{v}_i), \phi^*_i) \right| \leq L_\phi||\phi_i - \phi^*_i||, \\
    &\left| \mathcal{L}_i(h(\varphi_i; \mathbf{v}_i), \phi_i)  - \mathcal{L}_i(h^*(\varphi_i; \mathbf{v}_i), \phi_i) \right| \leq L_h||h_i - h^*_i||, \\
    &\left|| h(\varphi_i^*; \mathbf{v}_i)  - h(\varphi_i; \mathbf{v}_i)) \right|| \leq L_\varphi||\varphi_i - \varphi^*_i||, \\
    &\left|| h(\varphi_i; \mathbf{v}_i^*)  - h(\varphi_i; \mathbf{v}_i)) \right|| \leq L_\mathbf{v}||\mathbf{v}_i - \mathbf{v}^*_i||,
\end{aligned}
\end{equation*}
for $i=1, \cdots, m$.
\end{assumption}

\begin{theorem}
\label{theorem2}
Assuming we select $m$ clients during each FL communication round, let the hypernetwork parameter space be of dimension $H$, the embedding space be of dimension $d$, and the classifier parameter space be of dimension $K$. The parameters ${\phi}$, $\mathbf{v}$, ${\varphi}$  are those learned from each client's individual dataset. Under the condition of Assumption \ref{assumption_gen_bound}, we can derive:

\begin{equation} \label{eq:sample_size}
\begin{aligned}
S = &\mathcal{O}\bigg(\frac{d+H+K}{\epsilon^2}\log\Big(\frac{R(L_hL_\varphi+L_h L_v+ L_\phi)}{\epsilon}\Big)+\frac{1}{m\epsilon^2}\log\frac{1}{\delta}\bigg),
\end{aligned}
\end{equation}
such that if the number of samples per client exceeds $S$, then with a probability of at least $1 - \delta$, for all ${\phi}$, $\mathbf{v}$, ${\varphi}$, the following holds:
\begin{equation} 
| \sum_i^m \frac{n_i}{N}(\mathcal{L}_i(h(\varphi^*_i; \mathbf{v}_i^*), \phi^*_i) - \mathcal{L}_i(h(\varphi_i; \mathbf{v}_i), \phi_i))| \leq \epsilon, 
\end{equation}
where ${\phi^*}$, $\mathbf{v^*}$, ${\varphi^*}$ are the optimal parameters corresponding to the distribution of each individual client. 
\end{theorem}

Theorem \ref{theorem2} shows that the required sample size depends on the complexity of the parameter spaces and the number of clients. This highlights a trade-off between the generalization ability of the framework and the expressive power of the personalized models.

\section{Experiments}
\label{sec:exp}

In this section, we conduct a comprehensive experimental evaluation of the proposed HyperFL++ framework, assessing its effectiveness in terms of both privacy protection and model utility.

\subsection{Experimental Setup} \label{sec:exp_setup}

\subsubsection{Datasets} 
Our evaluation is performed on four widely-used image classification benchmarks: 
(1) \textbf{EMNIST} \cite{cohen2017emnist}, a dataset of handwritten characters across 62 classes.
(2) \textbf{Fashion-MNIST} \cite{xiao2017fashion}, comprising 10 categories of fashion product images.
(3) \textbf{CIFAR-10} \cite{krizhevsky2009learning}, a standard 10-class object recognition dataset.
(4) \textbf{CINIC-10} \cite{darlow2018cinic}, a larger 10-class dataset combining images from CIFAR-10 and ImageNet.

To simulate realistic federated learning scenarios, we follow the methodology in \cite{xu2023personalized, guo2025new} to create a non-IID data distribution. For each client, a small fraction of data ($s\%$, set to 20\%) is sampled uniformly from all classes, while the majority ($(100-s)\%$) is drawn from a limited set of dominant classes. Clients are organized into groups that share the same dominant classes. For Fashion-MNIST, CIFAR-10, and CINIC-10, we create 5 client groups, with each group assigned three consecutive classes (e.g., 0-2, 2-4, etc.). For EMNIST, we form 3 groups corresponding to digits, uppercase letters, and lowercase letters.

\subsubsection{Model Architectures} 
For the \textbf{Main Configuration of HyperFL++}, we adopt the CNN architectures from \cite{xu2023personalized, guo2025new}. A smaller CNN with two convolutional layers is used for EMNIST and Fashion-MNIST, while a larger CNN with three convolutional layers is used for CIFAR-10 and CINIC-10. The hypernetwork is a fully connected network with one hidden layer, generating the weights for the target CNNs. Client embeddings are learnable vectors of dimension 128.

For \textbf{HyperFL++-LPM}, we use a pre-trained ViT-S model \cite{dosovitskiy2021image} as the fixed feature extractor. Adapters, consisting of down-projection and up-projection layers with a ReLU activation, are inserted into each transformer block. The hypernetwork, with the same architecture as in the main configuration, generates the parameters for these adapters.

\subsubsection{Compared Methods}  
We benchmark HyperFL++ against a range of baselines. For the \textbf{Main Configuration of HyperFL++}, the baselines include:
(1) \textbf{FedAvg} \cite{mcmahan2017communication}, the standard FL algorithm.
(2) \textbf{pFedHN} and \textbf{pFedHN-HC} \cite{shamsian2021personalized}, which use a central hypernetwork for personalization.
(3) Several DP-based methods: \textbf{DP-FedAvg} \cite{mcmahan2018learning}, \textbf{PPSGD} \cite{bietti2022personalization}, and \textbf{CENTAUR} \cite{shen2023share}. Note that these methods were originally designed for user-level DP; we adapt them to the honest-but-curious server setting by applying DP protections before server aggregation.
(4) \textbf{HyperFL} \cite{guo2025new}, the precursor to our current work.

For \textbf{HyperFL++-LPM}, we compare against FedAvg with two fine-tuning strategies: (1) full parameter tuning and (2) adapter-based fine-tuning.

\subsubsection{Training Settings} 
For the \textbf{Main Configuration of HyperFL++}, we use mini-batch SGD as the local optimizer for all methods. The learning rates $\eta_h$ and $\eta_v$ are set to 0.01 for EMNIST/Fashion-MNIST and 0.02 for CIFAR-10/CINIC-10, while $\eta_g$ is 0.1 for all datasets. We use a weight decay of 5e-4, momentum of 0.5, and a batch size of 50 (100 for EMNIST). All experiments run for 200 global communication rounds with 5 local epochs. We test with both 20 clients (full participation) and 100 clients (30\% random sampling). For DP-based methods, we use a privacy budget of ($\epsilon=4, \sigma=1e-5$).

For \textbf{HyperFL++-LPM}, we use 20 clients and a batch size of 16. The learning rates $\eta_h$ and $\eta_v$ are 0.02 for EMNIST and 0.1 for CIFAR-10. 

All experiments are conducted on NVIDIA GeForce RTX 4090 and 3090 GPUs.

\subsection{Privacy Evaluation}
\label{sec:exp_privacy}

\begin{figure*}[t!]
    \centering
\includegraphics[width=1\textwidth]{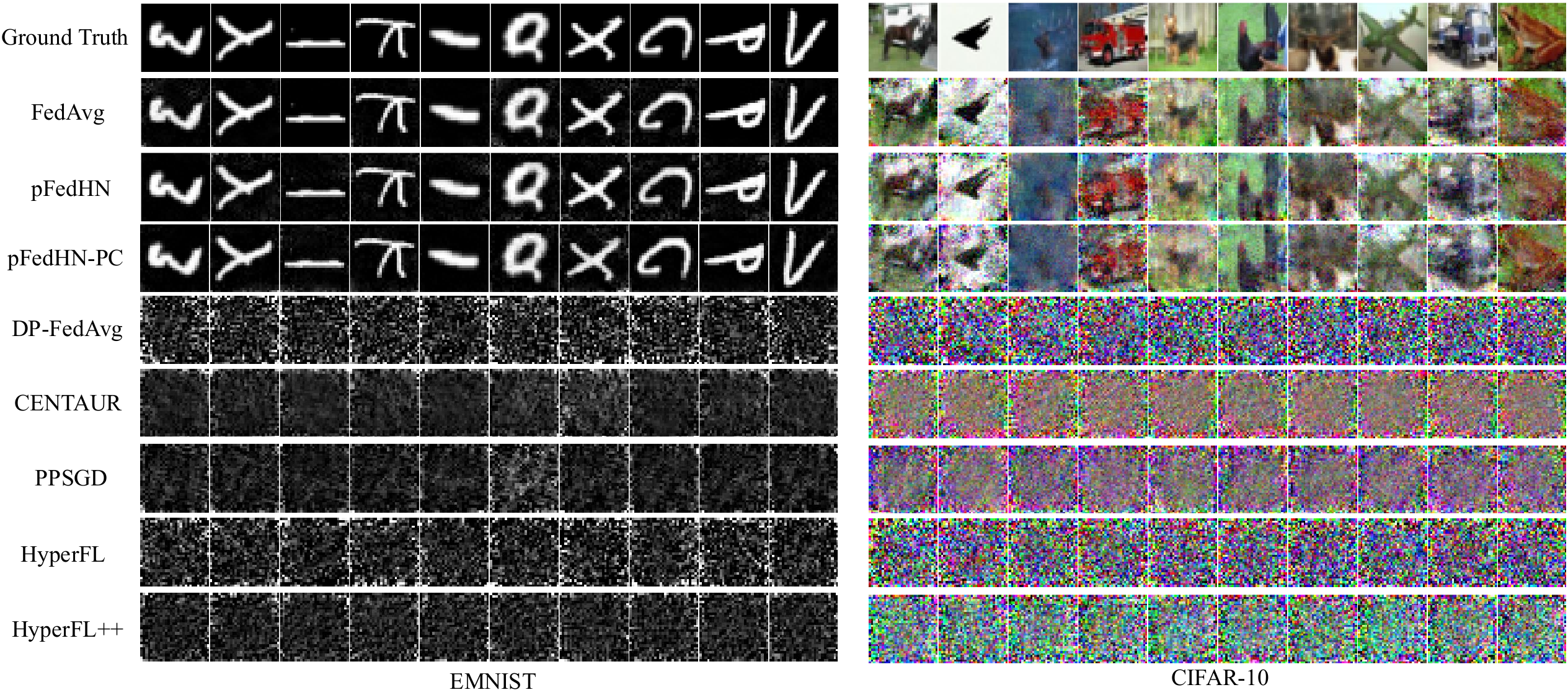}
    \caption{Visualization of reconstructed images of IG \cite{geiping2020inverting}. The images recovered from FedAvg, pFedHN, and pFedHN-PC are clearly recognizable, indicating severe privacy leakage. In contrast, images from DP-based methods, HyperFL, and HyperFL++ are noisy and unrecognizable, demonstrating effective privacy protection.}
    \label{fig:attack}
\end{figure*}

We experimentally validate the privacy-preserving capabilities of HyperFL++, complementing the theoretical analysis in Section \ref{sec:any_privacy}. We use the EMNIST and CIFAR-10 datasets for this evaluation, attempting to reconstruct a subset of 50 images (with a batch size of one) using the well-known IG attack method \cite{geiping2020inverting}. The attack is run for 10,000 optimization iterations. We measure the reconstruction quality using Peak Signal-to-Noise Ratio (PSNR), Structural Similarity (SSIM), and Learned Perceptual Image Patch Similarity (LPIPS). Better defense is indicated by lower PSNR/SSIM scores and a higher LPIPS score, as these signify a greater divergence between the reconstructed and original images.

The quantitative results are presented in Table \ref{tab:attack}, with corresponding visualizations in Figure \ref{fig:attack}. The results clearly expose the vulnerabilities of standard FL frameworks. FedAvg, pFedHN, and pFedHN-PC, which directly or indirectly expose local model information, are highly susceptible to GIA. This is evidenced by their high PSNR and SSIM scores and low LPIPS scores, which quantitatively confirm the high fidelity of the reconstructed images seen in Figure \ref{fig:attack}. While DP-based methods (DP-FedAvg, CENTAUR, PPSGD) offer strong privacy protection, as shown by their favorable metric scores, this comes at a steep cost to model performance (detailed in Table \ref{tab:main_result}).

\begin{table} [h]
\centering
\caption{Reconstruction results of IG \cite{geiping2020inverting}. Higher LPIPS, Lower PSNR, and SSIM of recovered images indicate better defense ability.}
\resizebox{0.7\linewidth}{!}{
  \begin{tabular}{l ccc ccc}
    \toprule
    \multirow{2}{*}{Method} & \multicolumn{3}{c}{EMNIST} & \multicolumn{3}{c}{CIFAR-10} \\
    \cmidrule(lr){2-4} \cmidrule(lr){5-7} 
     & PSNR  & SSIM  & LPIPS  & PSNR  & SSIM  & LPIPS  \\
    \midrule
    FedAvg \cite{mcmahan2017communication} & 32.64 & 0.8925 & 0.0526 & 16.16 & 0.6415 & 0.0536  \\
    pFedHN \cite{shamsian2021personalized} & 31.24 & 0.8701 & 0.0807 & 16.02 & 0.6351 & 0.0504 \\
    pFedHN-PC \cite{shamsian2021personalized} & 28.38 & 0.8713 & 0.0645 & 15.80 & 0.6247 & 0.0407 \\
    \midrule
    DP-FedAvg \cite{mcmahan2018learning} & 7.74 & 0.2978 & 0.7051 & 7.90 & 0.2716 & 0.3204  \\
    CENTAUR \cite{shen2023share} & 9.52 & 0.2136 & 0.6712 & 9.80 & 0.2723 & 0.2882  \\
    PPSGD \cite{bietti2022personalization} & 9.73 & 0.1889 & 0.6466 & 9.70 & 0.2788 & 0.2643 \\
    \midrule
    HyperFL \cite{guo2025new} & 7.85 & 0.3010 & 0.7147 & 8.35 & 0.2732 & 0.3132  \\
    \rowcolor{gray!30}
    HyperFL++ & 8.93 & 0.2274 & 0.6863 & 8.31 & 0.2796 & 0.7657 \\
    \bottomrule
  \end{tabular}
}
\label{tab:attack}
\end{table}

In stark contrast, both HyperFL and our proposed HyperFL++ achieve robust privacy protection, with metrics comparable to those of DP-based methods, while simultaneously delivering superior model accuracy (see Table \ref{tab:main_result}). This demonstrates the effectiveness of our core idea: ``breaking the direct connection between shared parameters and local data''.

\subsection{Utility Evaluation}

We now evaluate the utility of our proposed method in terms of model performance, system efficiency, and convergence rate.

\subsubsection{Model Performance}

As shown in Table \ref{tab:main_result}, the DP-based methods consistently underperform FedAvg. This is an expected outcome, as the injection of statistical noise, which is fundamental to DP, inherently degrades model utility to provide privacy guarantees. In contrast, our proposed HyperFL++ framework not only provides strong privacy but also achieves higher accuracy than the standard FedAvg in most tested scenarios. This result is significant as it demonstrates that HyperFL++ breaks the typical privacy-utility trade-off. While personalization methods like pFedHN and pFedHN-PC also show performance gains over FedAvg, they offer weak defense against GIA, as shown in our privacy evaluation, making them unsuitable for privacy-sensitive applications. HyperFL++ successfully combines the benefits of personalization (leading to higher accuracy) with robust privacy protection.

\begin{table*}[t]
\centering
\caption{The comparison of final test accuracy (\%) of different methods on various datasets. We apply full participation for FL system with 20 clients, and apply client sampling with rate 0.3 for FL system with 100 clients. \# Para [EMNIST/Fashion-MNIST; CIFAR-10/CINIC-10] denotes the communication parameters during each FL round. \underline{Underline} indicates better performance than FedAvg.}
\resizebox{1\linewidth}{!}{
\begin{tabular}{lccccccccc}
\toprule 
 \multirow{2}{*}{Method} & \multirow{2}{*}{\# Para} & \multicolumn{2}{c}{EMNIST} & \multicolumn{2}{c}{Fashion-MNIST} & \multicolumn{2}{c}{CIFAR-10} & \multicolumn{2}{c}{CINIC-10} \\
 \cmidrule(lr){3-4} \cmidrule(lr){5-6} \cmidrule(lr){7-8} \cmidrule(lr){9-10}
 & & 20 clients & 100 clients & 20 clients & 100 clients & 20 clients & 100 clients & 20 clients & 100 clients \\

\midrule
FedAvg \cite{mcmahan2017communication} & [115,776; 106,400] & 72.77 & 78.87 & 85.67 & 88.11 & 70.02 & 76.24 & 57.00 & 59.11 \\
\cmidrule(lr){1-10}
pFedHN \cite{shamsian2021personalized} & [115,776; 106,400] & \underline{80.86} & 77.37 & \underline{87.64} & \underline{89.80} & \underline{70.18} & \underline{80.07} & \underline{63.88} & \underline{70.36} \\
pFedHN-PC \cite{shamsian2021personalized} & [109,368; 110,168] & \underline{82.04} & \underline{79.32} & \underline{89.17} & \underline{90.06} & \underline{72.52} & \underline{78.15} & \underline{64.92} & \underline{68.73} \\
\cmidrule(lr){1-10}
DP-FedAvg \cite{mcmahan2018learning} & [115,776; 106,400] & 35.12 & 45.73 & 59.88 & 68.29 & 29.12 & 32.03 & 27.30 & 29.94 \\
CENTAUR \cite{shen2023share} & [115,776; 106,400] & 68.82	& 67.24	& 83.07	& 79.77 & 50.85 & 51.86& 48.82 & 51.01 \\
PPSGD \cite{bietti2022personalization} & [115,776; 106,400] & 71.16 & 71.18 & 84.47 &  82.94 & 52.17 & 53.92 & 49.98 & 52.91 \\
\cmidrule(lr){1-10}
HyperFL \cite{guo2025new} & [6,922,966; 6,924,256] & \underline{78.42} & \underline{80.71} & \underline{87.46} & \underline{89.38} & \underline{71.98} & {74.74} & \underline{66.52} & \underline{70.35} \\
\rowcolor{gray!30}
HyperFL++ & [70,136; 84,556] & \underline{78.18} & \underline{81.85} & \underline{88.26} & \underline{90.09} & \underline{71.80} & 74.76 & \underline{65.86} & \underline{69.52} \\
\bottomrule
\end{tabular}
}
\label{tab:main_result}
\end{table*}

Table \ref{tab:result_2} presents the results for HyperFL++-LPM, our adaptation for large pre-trained models. The results show that our method achieves performance comparable to FedAvg with full model fine-tuning, which is often considered the upper bound for performance but is computationally expensive and memory-intensive. More importantly, HyperFL++-LPM significantly outperforms FedAvg with adapter-only tuning. This is a crucial finding, as it demonstrates that our hypernetwork-based approach for generating adapter parameters is more effective than the standard method of simply averaging adapter parameters, confirming its effectiveness and utility in large-model settings.

\begin{table}[h]
\centering
\caption{The comparison of final test accuracy (\%) of different methods on various datasets with 20 clients. $^{\dagger}$ Adapter fine-tuning. $^{\dagger\dagger}$ Full parameter tuning.}
\resizebox{0.7\linewidth}{!}{
\begin{tabular}{lcccc}
\toprule 
Method & EMNIST & Fashion-MNIST & CIFAR-10 & CINIC-10 \\
\midrule


FedAvg$^{\dagger}$ \cite{mcmahan2017communication} & 75.86 & 83.95 & 68.55 & 65.90 \\
FedAvg$^{\dagger\dagger}$ \cite{mcmahan2017communication} & 72.73 & 85.69 & 92.37 & 91.37 \\
HyperFL++-LPM & 78.73 & 84.91 & 88.12 & 83.57 \\



\bottomrule
\end{tabular}
}
\label{tab:result_2}
\end{table}

\subsubsection{System Efficiency}

We assess system efficiency by measuring communication cost, per-round computation cost, and the number of rounds needed to reach a target accuracy (95\% of FedAvg's final accuracy), following the methodology in \cite{qu2022rethinking,guo2025selective}. As shown in Table \ref{tab:cost_emnist_cifar10}, HyperFL++ demonstrates remarkable efficiency. A key highlight is the drastic reduction in communication overhead compared to its predecessor, HyperFL. Thanks to the RKD technique, HyperFL++ reduces the much number of communicated parameters compared to HyperFL, bringing the cost down to a level even lower than that of standard FedAvg. While its per-round computation time is slightly higher than FedAvg's due to the hypernetwork's forward pass, this is more than compensated for by its superior convergence speed. HyperFL++ requires significantly fewer communication rounds to reach the target performance compared to all other baselines. For example, on CIFAR-10, it reaches the target in just 42 rounds, whereas FedAvg needs 123 rounds. This rapid convergence, combined with low communication costs, highlights the overall system efficiency of our framework, making it highly practical for real-world deployment.

\begin{table}[t]
\centering
\caption{Per-round communication and computation costs. \# Comm. Rounds denotes the number of rounds to reach the predefined target performance. - means it cannot reach the target performance.}
\resizebox{0.9\linewidth}{!}{
\begin{tabular}{lcccccc}
\toprule
\multirow{2}{*}{Method} & \multicolumn{2}{c}{\# Communicated Param} & \multicolumn{2}{c}{Computation Cost} & \multicolumn{2}{c}{\# Comm. Rounds} \\
\cmidrule(lr){2-3} \cmidrule(lr){4-5} \cmidrule(lr){6-7}
 & EMNIST & CIFAR-10 & EMNIST & CIFAR-10 & EMNIST & CIFAR-10 \\

\midrule
FedAvg \cite{mcmahan2017communication}     & 115,776 & 106,400 & 21.99 & 19.39 & 11 & 123 \\
\cmidrule(lr){1-7}
pFedHN \cite{shamsian2021personalized}     & 115,776 & 106,400 & 16.79 & 25.10 & 15 & 76 \\
pFedHN-PC \cite{shamsian2021personalized}  & 109,368 & 110,168 & 30.12 & 31.58 & 3 & 60 \\
\cmidrule(lr){1-7}
DP-FedAvg \cite{mcmahan2018learning}       & 115,776 & 106,400 & 202.52 & 144.33 & -- & -- \\
CENTAUR \cite{shen2023share}               & 115,776 & 106,400 & 187.47 & 140.39 & -- & -- \\
PPSGD \cite{bietti2022personalization}     & 115,776 & 106,400 & 225.99 & 177.11 & -- & -- \\
\cmidrule(lr){1-7}
HyperFL \cite{guo2025new}                  & 6,922,966 & 6,924,256 & 37.01 & 35.83 & 4 & 63 \\
\rowcolor{gray!30}
HyperFL++                                   & 70,136 & 84,556 & 47.83 & 40.04 & 3 & 42 \\
\bottomrule
\end{tabular}
}
\label{tab:cost_emnist_cifar10}
\end{table}

\subsubsection{Convergence Rate}

\begin{figure}[h]
  \centering
  \includegraphics[width=0.4\linewidth]{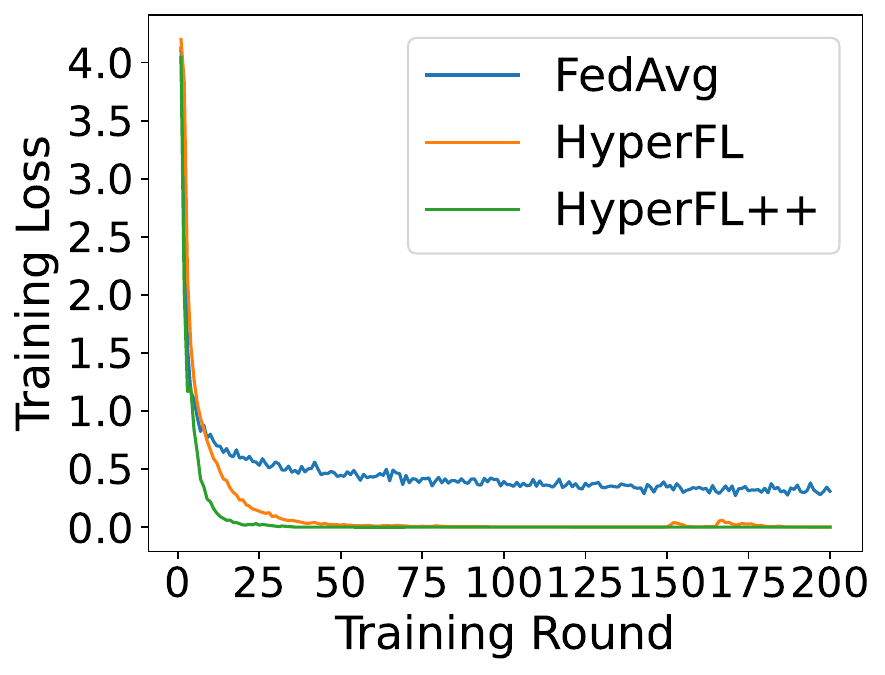} 
  \caption{
  Average training loss on the EMNIST dataset with 20 clients. The curve for HyperFL++ shows a smooth and rapid decrease, achieving a lower final loss than FedAvg, which indicates better model performance.
  }
  \label{fig:convergence}
\end{figure}

Figure \ref{fig:convergence} illustrates the training loss curves for FedAvg, HyperFL, and HyperFL++ on the EMNIST dataset. The plot shows that HyperFL++ not only converges smoothly but does so at a rate comparable to the standard FedAvg, validating the strong theoretical convergence properties established in Section \ref{sec:convergence_ana}. The training loss for HyperFL++ decreases steadily and reaches a low plateau, indicating stable and effective training. Furthermore, after convergence, HyperFL++ achieves a lower final training loss than FedAvg. This is a strong indicator of its superior modeling capacity, which directly correlates with the higher test accuracy reported in Table \ref{tab:main_result}. This demonstrates that the privacy-preserving mechanism of HyperFL++ does not hinder the optimization process; on the contrary, it facilitates better model fitting in heterogeneous data environments.

\section{Conclusion}
\label{sec:conclusion}

In this paper, we introduce \textbf{HyperFL++}, a novel federated learning framework designed to defend against Gradient Inversion Attacks by fundamentally altering the information-sharing mechanism. Instead of exchanging vulnerable model gradients, clients in our framework share only the parameters of a hypernetwork, which in turn generates the local model parameters. This approach effectively ``breaks the direct connection'' between shared updates and private data. To ensure the practicality of this design, we incorporate a Row-wise Kronecker Decomposition (\textbf{RKD}) technique, which significantly reduces the communication and memory overhead associated with the hypernetwork. Our extensive theoretical analysis and experimental results demonstrate that HyperFL++ provides robust privacy protection without compromising model utility or convergence speed. We hope this work will inspire further research into designing inherently private FL frameworks, shifting the focus from retrofitting defenses to building privacy-by-design systems.

\appendix

\section{Theoretical Proof} \label{sec:theoretical_proof}

\subsection{Proof of Theorem 1} \label{app_sec:proof_theo_1}

\begin{pf}
Let $\phi_i^t, \varphi_i^t, \mathbf{v}_i^t$ be the model parameters maintained in the $i$-th client at the $t$-th step. Let $\mathcal{I}_E$ be the set of global synchronization steps, i.e., $\mathcal{I}_E=\{nE \mid n=1,2,\cdots\}$. If $t+1 \in \mathcal{I}_E$, which represents the time step for communication, then the one-step update of HyperFL can be described as follows:
\begin{equation*}
\begin{array}{c}
\left(\begin{array}{c}
\phi_i^t \\
\varphi_i^t \\
\mathbf{v}_i^t
\end{array}\right) \underset{\text{SGD  of } \phi_i^t}{\longrightarrow}\left(\begin{array}{c}
\phi_i^{t+1} \\
\varphi_i^t \\
\mathbf{v}_i^t
\end{array}\right) \underset{\text{SGD of }\varphi_i^t, \mathbf{v}_i^t}{\longrightarrow}\left(\begin{array}{c}
\phi_i^{t+1} \\
\varphi_i^{t+1} \\
\mathbf{v}_i^{t+1}
\end{array}\right) 
\underset{\text { if } t+1 \in \mathcal{I}_E}{\longrightarrow}\left(\begin{array}{c}
\phi_i^{t+1} \\
\sum_{j=1}^m w_j \varphi_j^{t+1} \\
\mathbf{v}_i^{t+1}
\end{array}\right).
\end{array}
\end{equation*}
For convenience, we denote the parameters in each sub-step above as follows:
\begin{equation*}
\begin{array}{c}
x_i^t = \left(\begin{array}{c}
\phi_i^t \\
\varphi_i^t \\
\mathbf{v}_i^t
\end{array}\right),
y_i^t = \left(\begin{array}{c}
\phi_i^{t+1} \\
\varphi_i^{t} \\
\mathbf{v}_i^{t}
\end{array}\right),
x_i^{t+1,1} = \left(\begin{array}{c}
\phi_i^{t+1} \\
\varphi_i^{t+1} \\
\mathbf{v}_i^{t+1}
\end{array}\right),
x_i^{t+1,2} = \left(\begin{array}{c}
\phi_i^{t+1} \\
\sum_{j=1}^m w_j \varphi_j^{t+1} \\
\mathbf{v}_i^{t+1}
\end{array}\right),
\end{array}
\end{equation*}
\begin{equation*}
x_i^{t+1}= \begin{cases}x_i^{t+1, 1} & \text { if } t+1 \notin \mathcal{I}_E, \\ x_i^{t+1, 2} & \text { if } t+1 \in \mathcal{I}_E.\end{cases}
\end{equation*}
Here, the variable $x_i^{t+1, 1}$ represents the immediate result of one sub-step SGD update from the parameter of the previous sub-step $y_i^t$, and $x_i^{t+1, 2}$ represents the parameter obtained after communication steps (if possible). 
\\
Furthermore, we denote the learning rate and stochastic gradient of step $t$ as follows:
\begin{equation*}
\eta = \left(\begin{array}{c}
\eta_g \\
\eta_h \\
\eta_{v}
\end{array}\right),
\end{equation*}
\begin{equation*}
g_i^t=
\left(\begin{array}{c}
g_{i,\phi}^t \\
g_{i,\varphi}^t \\
g_{i,\mathbf{v}}^t
\end{array}\right)
= \left(\begin{array}{l}
\nabla_\phi \mathcal{L}_i\left(h\left(\mathbf{v}_i^t, \varphi_i^t\right), \phi_i^t, \xi_i^t\right) \\
\nabla_{\varphi} \mathcal{L}_i\left(h\left(\mathbf{v}_i^t, \varphi_i^t\right), \phi_i^{t+1}, \xi_i^t\right) \\
\nabla_\mathbf{v} \mathcal{L}_i\left(h\left(\mathbf{v}_i^t, \varphi_i^t\right), \phi_i^{t+1}, \xi_i^t\right)
\end{array}\right),
\end{equation*}
\begin{equation*}
\bar{g}_i^t=
\left(\begin{array}{c}
\bar{g}_{i,\phi}^t \\
\bar{g}_{i,\varphi}^t \\
\bar{g}_{i,\mathbf{v}}^t
\end{array}\right)
= \left(\begin{array}{l}
\nabla_\phi \mathcal{L}_i\left(h\left(\mathbf{v}_i^t, \varphi_i^t\right), \phi_i^t\right) \\
\nabla_{\varphi} \mathcal{L}_i\left(h\left(\mathbf{v}_i^t, \varphi_i^t\right), \phi_i^{t+1}\right) \\
\nabla_\mathbf{v} \mathcal{L}_i\left(h\left(\mathbf{v}_i^t, \varphi_i^t\right), \phi_i^{t+1}\right)
\end{array}\right),
\end{equation*}
where $\xi_i^t$ is the data uniformly chosen from the local data set of client $i$ at step $t$, then $\mathbb{E}\left[g_i^t\right] = \bar{g}_i^t$.
\\
\\
Next, we apply the inequality of the smoothness Assumption 1 to each sub-step of the one-step update for client $i$.
Firstly, by the smoothness of $\mathcal{L}_i$, we have:
\begin{equation}\label{l-smooth1}
\mathcal{L}_i\left(y_i^t\right) \leq \mathcal{L}_i\left(x_i^t\right)+\left\langle y_i^t-x_i^t, \bar{g}_{i,\phi}^t\right\rangle+\frac{L}{2}\left\|y_i^t-x_i^t\right\|^2.
\end{equation}
For the second term on the right side of inequality (\ref{l-smooth1}), according to the law of total expectation, we have:
\begin{equation*}
\begin{aligned} 
\mathbb{E}\left[\left\langle y_i^t-x_i^t, \bar{g}_{i,\phi}^t\right\rangle\right]
&= \mathbb{E}\left[\left\langle -\eta_g g_{i,\phi}^t, \bar{g}_{i,\phi}^t\right\rangle\right]\\  
&= \mathbb{E}\left\{\mathbb{E}\left[\left\langle -\eta_g g_{i,\phi}^t, \bar{g}_{i,\phi}^t\right\rangle\right]|\xi_i^t\right\}\\  
&= \mathbb{E}\left\{\mathbb{E}\left[\left\langle -\eta_g g_{i,\phi}^t|\xi_i^t, \bar{g}_{i,\phi}^t\right\rangle\right]\right\}\\  
&= \mathbb{E}\left[\left\langle -\eta_g \bar{g}_{i,\phi}^t, \bar{g}_{i,\phi}^t\right\rangle\right]\\
& = -\eta_{g}\mathbb{E}\left[(\bar{g}_{i,\phi}^t)^2\right].
\end{aligned} 
\end{equation*}
For the third term on the right side of the inequality (\ref{l-smooth1}), we have:
\begin{equation*}
\begin{aligned}
\mathbb{E}\left[\frac{L}{2}\left\|y_i^t-x_i^t\right\|^2\right]
&= \mathbb{E}\left[\frac{L}{2}\left\|-\eta_g g_{i,\phi}^t\right\|^2\right]\\
 &= \eta_g^2\frac{L}{2}\mathbb{E}\left[\left| g_{i,\phi}^t\right\|^2\right] \\
 &\leq \eta_g^2\frac{LG^2}{2},
\end{aligned}
\end{equation*}
where in the last inequality, we use the bounded gradient Assumption 3.\\
From the above inequalities, and taking the expectation of inequality (\ref{l-smooth1}), we can get:
\begin{equation}
\label{eq1}
\mathbb{E}\left[\mathcal{L}_i\left(y_i^t\right) - \mathcal{L}_i\left(x_i^t\right)\right] \leq -\eta_g \mathbb{E}\left[(\bar{g}_{i,\phi}^t)^2\right] + \eta_g^2\frac{LG^2}{2}.
\end{equation}
Secondly, by the smoothness of $\mathcal{L}_i$, we have:
\begin{equation}\label{l-smooth2}
\begin{aligned}
\mathcal{L}_i\left(x_i^{t+1,1}\right) \leq  \mathcal{L}_i\left(y_i^t\right)+\left\langle x_i^{t+1,1}-y_i^t, \bar{g}_{i,\phi}^t\right\rangle +\frac{L}{2}\left\|x_i^{t+1,1}-y_i^t\right\|^2.
\end{aligned}
\end{equation}
Similar to inequality (\ref{eq1}), taking the expectation of inequality (\ref{l-smooth2}), we get:

\begin{equation}
\label{eq2}
\begin{aligned}
\mathbb{E}\left[\mathcal{L}_i\left(x_i^{t+1,1}\right) - \mathcal{L}_i\left(y_i^t\right)\right] \leq -\eta_h\mathbb{E}\left[(\bar{g}_{i,\varphi}^t)^2\right] -\eta_{v}\mathbb{E}\left[(\bar{g}_{i,\mathbf{v}}^t)^2\right] + (\eta_h^2+\eta_{v}^2)\frac{LG^2}{2}.
\end{aligned}
\end{equation}
Thirdly, by the smoothness of $\mathcal{L}_i$, we have:

\begin{equation}\label{l-smooth3}
\begin{aligned}
\mathcal{L}_i\left(x_i^{t+1,2}\right) \leq \mathcal{L}_i\left(x_i^{t+1,1}\right) + \left\langle x_i^{t+1,2}-x_i^{t+1,1}, \bar{g}_{i,\varphi}^t\right\rangle +\frac{L}{2}\left\|x_i^{t+1,2}-x_i^{t+1,1}\right\|^2.
\end{aligned}
\end{equation}
From the iterative formula of SGD, it is clear that:

\begin{equation}
\label{cumm_update_equa}
    \varphi_j^{t+1} = \varphi_j^{t-E+1} - \eta_h \sum_{t_0=t-E+1}^tg_{j,\varphi}^{t_0}, \quad \forall j, t+1 \in \mathcal{I}_E.
\end{equation}
Then, for the third term on the right side of inequality (\ref{l-smooth3}), we apply the equality (\ref{cumm_update_equa}) and take the expectation, which yields:
\begin{equation*}
\begin{aligned}
\mathbb{E}\left[\frac{L}{2}\left\|x_i^{t+1,2}-x_i^{t+1,1}\right\|^2\right]
=  &\frac{L}{2} \mathbb{E}\left[\left\| - \eta_h\sum_{j=1}^m w_j \sum_{t_0=t-E+1}^t (g_{j,\varphi}^{t_0}-g_{i,\varphi}^{t_0}) \right\|^2\right] \\
\leq &\eta_h^2\frac{L}{2} \sum_{j=1}^m w_j\mathbb{E}\left[\left\| \sum_{t_0=t-E+1}^t (g_{j,\varphi}^{t_0}-g_{i,\varphi}^{t_0}) \right\|^2\right] \\
\leq& \eta_h^2\frac{L}{2} \sum_{j=1}^m w_j \sum_{t_0=t-E+1}^t \mathbb{E}\left[\left\| (g_{j,\varphi}^{t_0}-g_{i,\varphi}^{t_0}) \right\|^2\right]\\
\leq& \eta_h^2\frac{L}{2} \sum_{j=1}^m w_j \sum_{t_0=t-E+1}^t \mathbb{E}\left[\frac{1}{2} \left\|g_{j,\varphi}^{t_0}\right\|^2+\frac{1}{2} \left\|g_{i,\varphi}^{t_0}\right\|^2\right]\\
\leq& \eta_h^2 \frac{(E-1)LG^2}{2},
\end{aligned}
\end{equation*}
where in the last inequality, we use the bounded gradient Assumption 3.\\
For the second term on the right side of inequality (\ref{l-smooth3}), we take the expectation and get:
\begin{equation*}
\begin{aligned}
\mathbb{E}\left[\left\langle x_i^{t+1,2}-x_i^{t+1,1}, \bar{g}_{i,\varphi}^t\right\rangle\right] \leq & \frac{1}{2 \eta_h}\left(x_i^{t+1,2}-x_i^{t+1,1}\right)^2+\frac{1}{2} \eta_h \left(\bar{g}_{i, \varphi}^t\right)^2 \\
\leq & \frac{1}{2 \eta_h} \eta_h^2(E-1) G^2+\frac{1}{2} \eta_h \left(\bar{g}_{i, \varphi}^t\right)^2 \\
=& \eta_h \frac{(E-1) G^2}{2}+\frac{1}{2} \eta_h \left(\bar{g}_{i, \varphi}^t\right)^2,
\end{aligned}
\end{equation*}
where we use the Cauchy-Schwarz inequality and the AM-GM inequality in the first inequality, and the bounded gradient Assumption 3 in the second inequality above. \\
Then, based on the above inequalities and taking the expectation of inequality (\ref{l-smooth3}), we have:

\begin{equation}
\label{eq3}
\begin{aligned}
\mathbb{E}\left[\mathcal{L}_i\left(x_i^{t+1,2}\right) - \mathcal{L}_i\left(x_i^{t+1,1}\right)\right]
\leq \eta_h^2 \frac{(E-1)LG^2}{2} + \eta_h \frac{(E-1) G^2}{2}+\frac{1}{2} \eta_h \left(\bar{g}_{i, \varphi}^t\right)^2.
\end{aligned}
\end{equation}
Summing up inequalities (\ref{eq1}) and (\ref{eq2}), we get:

\begin{equation}
\begin{aligned}
\mathbb{E}\left[\mathcal{L}_i\left(x_i^{t+1,1}\right) - \mathcal{L}_i\left(x_i^t\right)\right] \leq -\eta^{\top} \mathbb{E}\left[(g_i^t)^2\right] + \left\|\eta\right\|^2 \frac{LG^2}{2}.
\end{aligned}
\end{equation}
Summing up inequalities (\ref{eq1}), (\ref{eq2}), and (\ref{eq3}), we get:

\begin{equation}
\begin{aligned}
\mathbb{E}\left[\mathcal{L}_i\left(x_i^{t+1,2}\right) - \mathcal{L}_i\left(x_i^t\right)\right]
\leq  -\left(\begin{array}{c}\eta_g \\ \frac{1}{2} \eta_h \\ \eta_v\end{array}\right)^{\top} \mathbb{E}\left[\left(\bar{g}_i^t\right)^2\right] + \left[\eta_g^2+\eta_v^2+(E-1) \eta_h^2+\frac{E-1}{L} \eta_h\right] \frac{L G^2}{2}.
\end{aligned}
\end{equation}
Then, let 
$\eta_{min} = \min\{\eta_g,\frac{1}{2}\eta_h,\eta_v\}$, we have:
\begin{equation} \label{neq:eta_min}
\begin{aligned}
\mathbb{E}\left[\mathcal{L}_i\left(x_i^{t+1,2}\right) - \mathcal{L}_i\left(x_i^t\right)\right] 
\leq -\eta_{min} \mathbb{E}\left[\left\|\bar{g}_i^t\right\|^2\right] +\left[\eta_g^2+\eta_v^2+(E-1) \eta_h^2+\frac{E-1}{L} \eta_h\right] \frac{L G^2}{2}.
\end{aligned}
\end{equation}
By rewriting the above inequality (\ref{neq:eta_min}), we get:

\begin{equation} \label{neq:e_g}
\begin{aligned}
\mathbb{E}\left[\left\|\bar{g}_i^t\right\|^2\right] \leq \frac{\mathbb{E}\left[\mathcal{L}_i\left(x_i^{t}\right) - \mathcal{L}_i\left(x_i^{t+1}\right)\right]}{\eta_{min}} + \frac{\eta_g^2+\eta_v^2+(E-1) \eta_h^2+\frac{E-1}{L} \eta_h}{\eta_{min}}\frac{LG^2}{2}.
\end{aligned}
\end{equation}
Let $M$ be a constant that satisfies the inequality $\eta_g^2+\eta_v^2+(E-1) \eta_h^2+\frac{E-1}{L} \eta_h \leq M \eta_{min}^2$, the aforementioned inequality (\ref{neq:e_g}) can be further simplified as:

\begin{equation}
\label{leq1}
\mathbb{E}\left[\left\|\bar{g}_i^t\right\|^2\right] \leq \frac{\mathbb{E}\left[\mathcal{L}_i\left(x_i^{t}\right) - \mathcal{L}_i\left(x_i^{t+1}\right)\right]}{\eta_{min}} + \eta_{min} \frac{LMG^2}{2}.
\end{equation}
Now, by repeatedly applying inequality (\ref{leq1}) for different values of $t$ and summing up the results, we get:
\begin{equation}
\label{leq29}
\begin{aligned}
\sum_{t=1}^{T} \mathbb{E}\left[\left\|\bar{g}_i^t\right\|^2\right] \leq \frac{\mathbb{E}\left[\mathcal{L}_i\left(x_i^{1}\right) - \mathcal{L}_i\left(x_i^*\right)\right]}{\eta_{min}} + \eta_{min} \frac{LMG^2}{2}T.
\end{aligned}
\end{equation}
Dividing both side of inequality (\ref{leq29}) by $T$, we get:
\begin{equation}
\begin{aligned}
\frac{1}{T}\sum_{t=1}^{T} \mathbb{E}\left[\left\|\bar{g}_i^t\right\|^2\right] \leq \frac{\mathbb{E}\left[\mathcal{L}_i\left(x_i^{1}\right) - \mathcal{L}_i\left(x_i^*\right)\right]}{\eta_{min}T}+ \eta_{min} \frac{LMG^2}{2}.
\end{aligned}
\end{equation}
Let us assume that $\mathcal{L}_i\left(x_i^{1}\right) - \mathcal{L}_i\left(x_i^*\right) \leq D, \forall i$, and we set $\eta_{min} = \sqrt{\frac{2D}{LMG^2T}}$. Then, we have:
\begin{equation}
\frac{1}{T}\sum_{t=1}^{T} \mathbb{E}\left[\left\|\bar{g}_i^t\right\|^2\right] \leq 2\sqrt{\frac{LMG^2D}{2T}}.
\end{equation}
Thus, we can get:
\begin{equation}
\frac{1}{mT}\sum_{i=1}^{m}\sum_{t=1}^{T} \mathbb{E}\left[\left\|\bar{g}_i^t\right\|^2\right] \leq 2\sqrt{\frac{LMG^2D}{2T}}.
\end{equation}
\end{pf}

\subsection{Proof of Corollary 1}
\begin{pf}
By rewriting inequality (\ref{leq1}), we have:
\begin{equation}
\label{leq32}
\begin{aligned}
\mathbb{E}\left[\mathcal{L}_i\left(x_i^{t+1}\right)-\mathcal{L}_i\left(x_i^t\right)\right] 
\leq -\eta_{min } \mathbb{E}\left[\left\|\bar{g}_i^t\right\|^2\right] +\eta_{min }^2 \frac{L M G^2}{2}.
\end{aligned}
\end{equation}
By the PL Assumption 4, we have:

\begin{equation}
\label{leq33}
\begin{aligned}
\mathbb{E}\left[\mathcal{L}_i\left(x_i^{t+1}\right)-\mathcal{L}_i\left(x_i^t\right)\right]
\leq -2\eta_{min}\mu \mathbb{E}\left[\mathcal{L}_i\left(x_i^{t}\right) - \mathcal{L}_i\left(x_i^*\right)\right] + \eta_{\min }^2 \frac{L M G^2}{2}.
\end{aligned}
\end{equation}
Then, 
{\small
\begin{equation*} 
\begin{aligned}
\mathbb{E}\left[\mathcal{L}_i\left(x_i^{t+1}\right) - \mathcal{L}_i\left(x_i^*\right)\right]= & \mathbb{E}\left[\mathcal{L}_i\left(x_i^{t}\right)- \mathcal{L}_i\left(x_i^*\right)\right] + \mathbb{E}\left[\mathcal{L}_i\left(x_i^{t+1}\right)-\mathcal{L}_i\left(x_i^t\right)\right]\\
\leq & \mathbb{E}\left[\mathcal{L}_i\left(x_i^{t}\right)- \mathcal{L}_i\left(x_i^*\right)\right]-2\eta_{min}\mu \mathbb{E}\left[\mathcal{L}_i\left(x_i^{t}\right) - \mathcal{L}_i\left(x_i^*\right)\right] + \eta_{min}^2\frac{LMG^2}{2}\\
= & (1-2\eta_{min}\mu) \mathbb{E}\left[\mathcal{L}_i\left(x_i^{t}\right) - \mathcal{L}_i\left(x_i^*\right)\right] + \eta_{min}^2\frac{LMG^2}{2}\\
\leq & \left(1-2 \eta_{min } \mu\right)^2 \mathbb{E}\left[\mathcal{L}_i\left(x_i^{t-1}\right)-\mathcal{L}_i\left(x_i^*\right)\right] +\sum_{\tau=0}^1\left(1-2 \eta_{\min } \mu\right)^\tau \eta_{\min }^2 \frac{L M G^2}{2} \\
\cdots \\
\leq & \left(1-2 \eta_{min } \mu\right)^{t+1} \mathbb{E}\left[\mathcal{L}_i\left(x_i^0\right)-\mathcal{L}_i\left(x_i^*\right)\right]+\sum_{\tau=0}^t\left(1-2 \eta_{min } \mu\right)^\tau \eta_{min }^2 \frac{L M G^2}{2}\\
\leq & (1-\eta_{min}\mu)^{t+1} D + \eta_{min}\frac{LMG^2}{4\mu}.
\end{aligned}
\end{equation*}
}
Therefore, we have:
\begin{equation}
\begin{aligned}
\mathbb{E}\left[\frac{1}{m}\sum_{i=1}^m\mathcal{L}_i\left(x_i^{t+1}\right)\right] - \mathcal{L}^* 
\leq (1-\eta_{min}\mu)^{t+1} D+ \eta_{min}\frac{LMG^2}{4\mu}.
\end{aligned}
\end{equation}
If we set $\eta_{min} \leq \frac{\mu\epsilon}{LMG^2}$, after $O(\frac{1}{\epsilon} \log (\frac{1}{\epsilon}))$ steps, we have:
\begin{equation}
\mathbb{E}\left[\frac{1}{m}\sum_{i=1}^m\mathcal{L}_i\left(x_i^{t+1}\right)\right] - \mathcal{L}^* \leq \epsilon.
\end{equation}
\end{pf}

\subsection{Proof of Theorem 2}
\begin{pf}    
First, we define the distance between ($\mathbf{\phi}$, $\mathbf{v}$, $\mathbf{\varphi}$) and ($\mathbf{\phi^*}$, $\mathbf{v^*}$, $\mathbf{\varphi^*}$) as:
\begin{equation}
\begin{aligned}
d\big((\mathbf{\phi}, \mathbf{v}, \mathbf{\varphi}),(\mathbf{\phi^*}, \mathbf{v^*}, \mathbf{\varphi^*})\big) 
= \Big| \sum_i^m \frac{n_i}{N}(\mathcal{L}_i(h(\varphi^*_i; \mathbf{v}_i^*), \phi^*_i) - \mathcal{L}_i(h(\varphi_i; \mathbf{v}_i), \phi_i))\Big|,
\end{aligned}
\end{equation}
where $N = \sum_{i=1}^m n_i$.
By the Theorem 4 from \cite{baxter2000model}, we can find an $\epsilon$-covering in $d((\mathbf{\phi}, \mathbf{v}, \mathbf{\varphi}),(\mathbf{\phi^*}, \mathbf{v^*}, \mathbf{\varphi^*}))$. Then, according to the notations used in our paper, we have that $S = \mathcal{O}(\frac{1}{m\epsilon^2}log(\frac{\mathcal{C}(\epsilon, \mathcal{H}_l^n)}{\delta}))$, where $\mathcal{C}(\epsilon, \mathcal{H}_l^n)$ is the covering number of $\mathcal{H}_l^n$. In our case, each element of $\mathcal{H}_l^n$ is parameterized by $\mathbf{\phi}, \mathbf{v}, \mathbf{\varphi}$. Therefore, from the triangle inequality and the Lipschitz conditions in Assumption 3, we can get: 

{\small
\begin{equation}
\begin{aligned}
d\big((\mathbf{\phi}, \mathbf{v}, \mathbf{\varphi}),(\mathbf{\phi^*}, \mathbf{v^*}, \mathbf{\varphi^*})\big)
= &\big|\sum_i^m\frac{n_i}{N}(\mathcal{L}_i(h(\varphi^*_i; \mathbf{v}_i^*), \phi^*_i) -  \mathcal{L}_i(h(\varphi_i; \mathbf{v}_i), \phi_i))\big| \\
= &\big|\sum_i^m\frac{n_i}{N}(\mathcal{L}_i(h(\varphi^*_i; \mathbf{v}_i^*), \phi^*_i) -  \mathcal{L}_i(h(\varphi_i; \mathbf{v}_i^*), \phi_i^*) \\
&+ \mathcal{L}_i(h(\varphi_i; \mathbf{v}_i^*), \phi_i^*) -  \mathcal{L}_i(h(\varphi_i; \mathbf{v}_i), \phi_i^*) + \mathcal{L}_i(h(\varphi_i; \mathbf{v}_i), \phi_i^*) -  \mathcal{L}_i(h(\varphi_i; \mathbf{v}_i), \phi_i))\big| \\
\leq &\sum_i^m \frac{n_i}{N}\big| (\mathcal{L}_i(h(\varphi^*_i; \mathbf{v}_i^*), \phi^*_i) - \mathcal{L}_i(h(\varphi_i; \mathbf{v}_i^*), \phi_i^*) \\ 
&+  \mathcal{L}_i(h(\varphi_i; \mathbf{v}_i^*), \phi_i^*) -  \mathcal{L}_i(h(\varphi_i; \mathbf{v}_i), \phi_i^*) +  \mathcal{L}_i(h(\varphi_i; \mathbf{v}_i), \phi_i^*) -  \mathcal{L}_i(h(\varphi_i; \mathbf{v}_i), \phi_i))\big| \\
\leq &\sum_{i=1}^m\frac{n_i}{N} (L_h||h(\varphi_i^*; \mathbf{v}_i^*)-h(\varphi_i; \mathbf{v}_i^*)||+L_h||h(\varphi_i; \mathbf{v}_i^*) - h(\varphi_i; \mathbf{v}_i)||+L_\phi||\phi_i^* - \phi_i||) \\
\leq & L_h L_\varphi||\varphi_i^*-\varphi_i|| + L_h L_v||\mathbf{v}_i^* - \mathbf{v}_i||+L_\phi||\phi_i^* - \phi_i||.
\end{aligned}
\end{equation}
}

\noindent Now if there is a parameter space such that $\phi_i$, $\mathbf{v}_i$ and $\varphi_i$ have corresponding optimal point $\phi_i^*$, $\mathbf{v}_i^*$ and $\varphi_i^*$, which are $\frac{\epsilon}{L_h L_\varphi+ L_h L_v+L_\phi}$ away, respectively, we can get an upper bound of the distance between our model and optimal model, which is an $\epsilon$-covering in $d((\mathbf{\phi}, \mathbf{v}, \mathbf{\varphi}),(\mathbf{\phi^*}, \mathbf{v^*}, \mathbf{\varphi^*}))$ matrix. From here we see that $\log(\mathcal{C}(\epsilon, \mathcal{H}_l^n)) = \mathcal{O}(m(d+H+K)\log(\frac{RL_h(L_\varphi + L_v)+R L_\phi}{\epsilon}))$.  
\end{pf}

\section*{Acknowledgments}

This work was supported by National Natural Science Foundation of China (62306253, 62206075), Guangdong Natural Science Fund-General Programme (2024A1515010233), Research Grants Council of the Hong Kong SAR, China (27207025), and UCSC hellman fellowship.

\bibliographystyle{elsarticle-num} 
\bibliography{PR/ref}






\end{document}